\documentclass{article}
\usepackage{spconf,amsmath,epsfig}
\usepackage{hyperref}
\usepackage{amsmath}
\usepackage{cite}
\usepackage{amssymb}
\usepackage{algorithm}
\usepackage{algpseudocode}
\usepackage{balance}
\usepackage{hyperref}
\usepackage{bbm}
\usepackage[table,xcdraw]{xcolor}
\usepackage{color}
\usepackage{soul}
\usepackage{subfig}
\usepackage{wrapfig}

\hypersetup{
    bookmarks=true,         
    unicode=false,          
    pdftoolbar=true,        
    pdfmenubar=true,        
    pdffitwindow=false,     
    pdfstartview={FitH},    
    pdftitle={sparse DTM},    
    pdfauthor={Camps-Valls},     
    pdfsubject={sparse DTM},   
    pdfcreator={Camps-Valls},   
    pdfproducer={Camps-Valls}, 
    pdfkeywords={keyword1} {key2} {key3}, 
    pdfnewwindow=true,      
    colorlinks=true,       
    linkcolor=blue,          
    citecolor=blue,        
    filecolor=blue,      
    urlcolor=blue           
}
\def\x{{\mathbf x}}
\def\z{{\mathbf z}}
\def\Z{{\mathbf Z}}
\def\X{{\mathbf X}}
\def\K{{\mathbf K}}
\def\bmu{{\boldsymbol \mu}}
\def\bSigma{{\boldsymbol \Sigma}}

\def\Real{{\mathbbm{R}}}
\def\Comp{{\mathbbm{C}}}
\newcommand{\imag}{\mathbbm{i}}

\title{Randomized RX for target detection}

\name{Fatih Nar$^{1}$, Adri\'an P\'erez-Suay$^{2}$, Jos\'e Antonio Padr\'on$^{2}$, Gustau Camps-Valls$^{2}$
	\thanks{This work was supported by the Scientific and Technical Research Council of Turkey (TUBITAK), Grant Number: TUBITAK-BIDEB-2219. GCV was funded by the European Research Council (ERC) under the ERC-CoG-2014 SEDAL project (grant agreement 647423), and APS was funded by the Spanish Ministry of Economy and Competitiveness (MINECO) and European Regional Development Fund (ERDF) through the project TIN2015-64210-R.}
}
\address{$^{1}$Konya Food and Agriculture University, Konya, Turkey \\
$^{2}$Image Processing Lab (IPL), Universitat de Val\`encia, Val\`encia, Spain}

\begin{document}

\maketitle
\begin{abstract}

This work tackles the target detection problem through the well-known global RX method.
The RX method models the clutter as a multivariate Gaussian distribution, and has been extended to nonlinear distributions using kernel methods. 
While the kernel RX can cope with complex clutters, it requires a considerable amount of computational resources as the number of clutter pixels gets larger.
Here we propose random Fourier features to approximate the Gaussian kernel in kernel RX and consequently our development keep the accuracy of the nonlinearity while reducing the computational cost which is now controlled by an hyperparameter. 
Results over both synthetic and real-world image target detection problems show space and time efficiency of the proposed method while providing high detection performance.

\end{abstract}


\section{Introduction}
\label{sec:introduction}

Anomaly Detection (AD) in remote sensing data analysis has been (and will continue to be) one of the main important research topics for various applications~\cite{Stein02,Keshava04}. 
Anomaly decision is made by analyzing the difference of the pixel under test (PUT) and the background.
Various anomaly detectors have been proposed in the literature under a number of assumptions: 1) the choice of the background region and the distribution, 2) the way of calculating the distance between the PUT and the background statistics, and 3) the definition of the decision threshold or test statistic \cite{Matteoli2010}. 

Among the large variety of AD methods, the Reed-Xiaoli (RX) is widely used for its simplicity, good practical results and efficiency in operational settings~\cite{Stein02,Keshava04}. 
The RX algorithm, which probably is the most known AD method for hyperspectral images, it is based on assuming a multivariate Gaussian distributed background and to deal with, it uses the Mahalanobis distance between the PUT and the background~\cite{Reed90tassp}. 
With regard to the background region, there are two main different approaches. In the global approach the whole image is defined as the background while in the local approach pixels around the PUT are defined as the background by means of a sliding window or a segmentation based approach~\cite{Molero13jstars,Zhao15rs}. 
Also, possible anomalies can be censored to prevent degradation on the background statistics~\cite{Nar17dsp}. 

Even though RX is computationally efficient and works well for simple backgrounds (mostly linear), it can not cope with more complex, nonlinear and\slash or non-Gaussian backgrounds.
To deal with its limitations, more powerful techniques like kernel methods can be used to cope with non-Gaussian distributed data~\cite{CampsValls09wiley,Kwon07,Longbotham14}. 
However, kernel RX is a computationally demanding method because it involves matrix inversion in the size of the sample count, e.g. for more than thousand pixels.
On the contrary, covariance matrix in RX has fixed size which is proportional to spectral dimension of the image, e.g. around ten for multispectral images and around hundred for hyperspectral images. 
Since AD is generally the first step in further recognition tasks, the execution time is as (or even more) important as the detection accuracy \cite{Chen14taes}.
Besides, processor capacity (CPU or GPU), memory storage, and execution budget can be limited in operational applications \cite{Molero13jstars}. 
In \cite{Zhao16rs}, a fast recursive kernel RX method was recently introduced, which processes the data in a causal manner. 
Although this method is fast, threshold estimation for target decision is a problematic task since the detection result is dependent on the location of the PUT in the image.

In this study, we propose an alternative way to speed up the kernel RX method while maintaining its accuracy: we propose to approximate the kernel function with random Fourier features~\cite{Rahimi07nips} in order to project data into a nonlinear feature space where a standard, cheap and efficient linear RX method can be readily applied. 
Previous use of the randomized approach in remote sensing has considered classification and regression problems~\cite{Valero15igarss,Perez17rks}. 
The proposed randomized RX approach is able to significantly lower the execution time and the memory storage compared to its kernel RX counterpart, while preserving almost the same detection accuracy. 
We give empirical evidence in both synthetic and real-world remote sensing target detection problems.

The remainder of the paper is organized as follows.
First, section \ref{sec:randomized_GRX} briefly reviews the proposed method used in this study. 
Then, section \ref{sec:experiments} presents the performance of the randomized RX method.
Finally, we conclude in section \ref{sec:conclusion} with some remarks and prospective future work.

\section{Randomized RX}
\label{sec:randomized_GRX}

\subsection{The RX algorithm}
\label{sec:RX_algorithm}

Let us define a hyperspectral image in matrix form $\X\in\mathbbm{R}^{n\times d}$, where $n$ is the number of pixels and $d$ is the dimensionality of each pixel, i.e. number of spectral channels. 
A generic pixel element in $I$ is denoted as the (column) feature vector $\x_i\in\mathbbm{R}^{d}$.

Among the various AD proposed in the literature, one of the most frequently used anomaly detector is the (spectral only version of the) Reed-Xiaoli (RX) detector~\cite{Reed90tassp} that is often used as a benchmark to which other methods are compared. 
The RX detector characterizes the background by its spectral mean vector $\bmu$ and covariance matrix $\bSigma_{RX} = \frac{1}{d}\tilde{\X}^\top\tilde{\X}$, where $\tilde{\X}$ is the centered $\X$ matrix. 
The actual detector calculates the Mahalanobis distance between the pixel under test, $\x_\ast$, and the background as follows
\begin{equation}\label{eq:rx}
D_{RX}(\x_\ast)=(\x_\ast-\bmu)^{\top}\bSigma_{RX}^{-1}(\x_\ast-\bmu).
\end{equation}
Note that we only need to invert the covariance matrix once. 
However, the global RX detector characterizes the background of the complete scene by a single multivariate normal probability density function (pdf). 
In many scenes, this model is not adequate. 
For this reason, several variations of the global RX detector have been proposed using kernel functions, such as the kernel RX (KRX)~\cite{Kwon07}.

\subsection{The kernel RX algorithm}

Notationally, let us map all pixels to a higher dimensional Hilbert feature spaces ${\mathcal H}$ by means of the feature map $\boldsymbol{\phi}: \x\to\boldsymbol{\phi}(\x)$. 
The mapped training data matrix $\X\in\Real^{n\times d}$ is now denoted as $\boldsymbol{\Phi}\in\Real^{n \times d_{\mathcal H}}$. 
Let us define a kernel function $K$ that, by virtue of the Riesz theorem, can evaluate (reproduce) the dot product between samples in ${\mathcal H}$, i.e. $K(\x,\x')=\langle\boldsymbol{\phi}(\x),\boldsymbol{\phi}(\x')\rangle\in{\mathbb R}$. 
Now, in order to estimate the anomalousness (distance) for a test example $\x_\ast$, we first map it $\boldsymbol{\phi}(\x_\ast)$ and then apply the RX formula in~\eqref{eq:rx}:
$$D_{KRX}(\x_\ast) = \boldsymbol{\phi}(\x_\ast)^\top(\boldsymbol{\Phi}^\top\boldsymbol{\Phi})^{-1}\boldsymbol{\phi}(\x_\ast),$$ 
which, after some linear algebra, can be expressed in terms of kernel matrices~\cite{ShaweTaylor04,CampsValls09wiley}:
\begin{equation}
\label{eq:krx}
D_{KRX}(\x_\ast) = {\bf k}_\ast^\top(\K\K)^{-1}{\bf k}_\ast,
\end{equation} 
where ${\bf k}_\ast=[K(\x_\ast,\x_1),\ldots,K(\x_\ast,\x_n)]^\top\in\Real^{n}$ contains the similarities between $\x_\ast$ and all training points in $\X$ using $K$, and $\K\in\Real^{n\times n}$ stands for the kernel matrix containing all training data similarities. 
Note that, if $n$ is very large constructing and inverting kernel matrix is not feasible hence generally $N$ samples are randomly sampled to obtain computational tractability.
Recently, the kernel RX has been also extended to non-Gaussian settings by defining a complete family of anomalous change detectors with kernels~\cite{Longbotham14}. 

\subsection{The randomized RX algorithm}
\label{sec:randomized_kernels}

An outstanding result in the recent kernel methods literature makes use of a classical definition in harmonic analysis to improve approximation and scalability~\cite{Rahimi07nips}. 
The Bochner's theorem states that a continuous shift-invariant kernel $K(\x,\x')=K(\x-\x')$ on $\Real^d$ is positive definite (p.d.) if and only if $K$ is the Fourier transform of a non-negative measure. 
If a shift-invariant kernel $K$ is properly scaled, its Fourier transform $p({\bf w})$ is a proper probability distribution. 
This property is used to approximate kernel functions with linear projections on a number of $D$ random features, as follows:
$$
K(\x, \x')  
\approx  \dfrac{1}{D} \sum\nolimits_{j=1}^D \exp(-\imag{\bf w}_j^\top\x)\exp(\imag{\bf w}_j^\top\x'), 
$$
where $\imag=\sqrt{-1}$, and ${\bf w}_i \in \Real^{d}$ is randomly sampled from a data-independent distribution $p({\bf w})$~\cite{Rahimi08}. 
Note that we can define a $D$-dimensional {\em randomized} feature map ${\bf z}: \Real^d\to\Comp^D$, which can be {\em explicitly} constructed as ${\bf z}({\bf x}):=[\exp(\imag{\bf w}_1^\top{\bf x}),\ldots,\exp(\imag{\bf w}_D^\top{\bf x})]^\top$. 

Therefore, given $n$ data points (pixels), the kernel matrix ${\bf K} \in\Real^{n\times n}$ can be approximated with the explicitly mapped data, ${\bf Z}=[{\bf z}_1\cdots{\bf z}_n]^\top\in\Real^{n\times D}$, and will be denoted as $\hat{\bf K}\approx{\bf Z}{\bf Z}^\top.$ 
However, we do not use such approximation in Eq.~\eqref{eq:krx} which would lead to a mere approximation with extra computational cost. 
Instead, we run the linear RX in Eq.~\eqref{eq:rx} with explicitly mapped points onto random Fourier features, which reduces to
\begin{equation}
D_{RRX} = \Re{\left(\z_\ast^\top(\Z^\top\Z)^{-1}\z_\ast\right)},
\end{equation}
where $\z_\ast = \z(\x_\ast)$, and $\Re$ is the real part function $\Re{\left(a+\imag b\right)}=a$. 
This leads to a nonlinear randomized RX (RRX) that approximates the KRX in expectation. 
Essentially, we map the original data $\x_i$ into a nonlinear space through the explicit mapping $\z(\x_i)$ to a $D$-dimensional space (instead of the potentially infinite feature space with $\boldsymbol{\phi}(\x_i)$), and then use the linear RX formula.
This allows to control the space and time complexity explicitly through $D$, as one has to store matrices of $n\times D$ and invert matrices of size $D\times D$ only (see Table~\ref{spacetime_complexity}).
Typically, parameter $D$ satisfies $D\ll n$ in practical applications, which turns to be a beneficial regularization effect in the solution.

\begin{table}[h]
\begin{center}
\captionsetup{skip=1pt}
\caption{Space and time complexity for all methods.}
\label{spacetime_complexity}
\begin{tabular}{|l|l|l|| l|l|l|l|}
\hline
\rowcolor[gray]{.60}
& \multicolumn{2}{c||}{Space}     & \multicolumn{4}{c}{Time}       \\ \hline
\rowcolor[gray]{.90}
Method & $T$     & $C^{-1}$   & $T$      & $C$      & $C^{-1}$   & $AD$      \\ \hline \hline
RX     & $-$     & $d^2$      & $-$      & $nd^2$   & $d^3$      & $nd^2$    \\ \hline
KRX    & $nN$    & $N^2$      & $ndN$    & $N^3$    & $N^3$      & $nN^2$    \\ \hline
RRX    & $nD$    & $D^2$      & $ndD$    & $nD^2$   & $D^3$      & $nD^2$    \\ \hline
\end{tabular}
\vspace*{1pt}
\newline\footnotesize{$T$ is transformation of image into a nonlinear space.}
\newline\footnotesize{$C$ is covariance matrix and $C^{-1}$ is its inverse.}
\newline\footnotesize{$AD$ is anomaly detection.}
\end{center}
\end{table}

\section{Experiments}
\label{sec:experiments}

This section reports some empirical results comparing RX, the KRX and the proposed RRX, in both synthetic and real experiments which are run on Intel i7 4.0 GHz PC.

\subsection{Synthetic dataset}
\label{sec:synthetic_data_experiment}

\begin{wrapfigure}{l}{0.60\columnwidth}
\centering{\includegraphics[width=0.60\columnwidth]{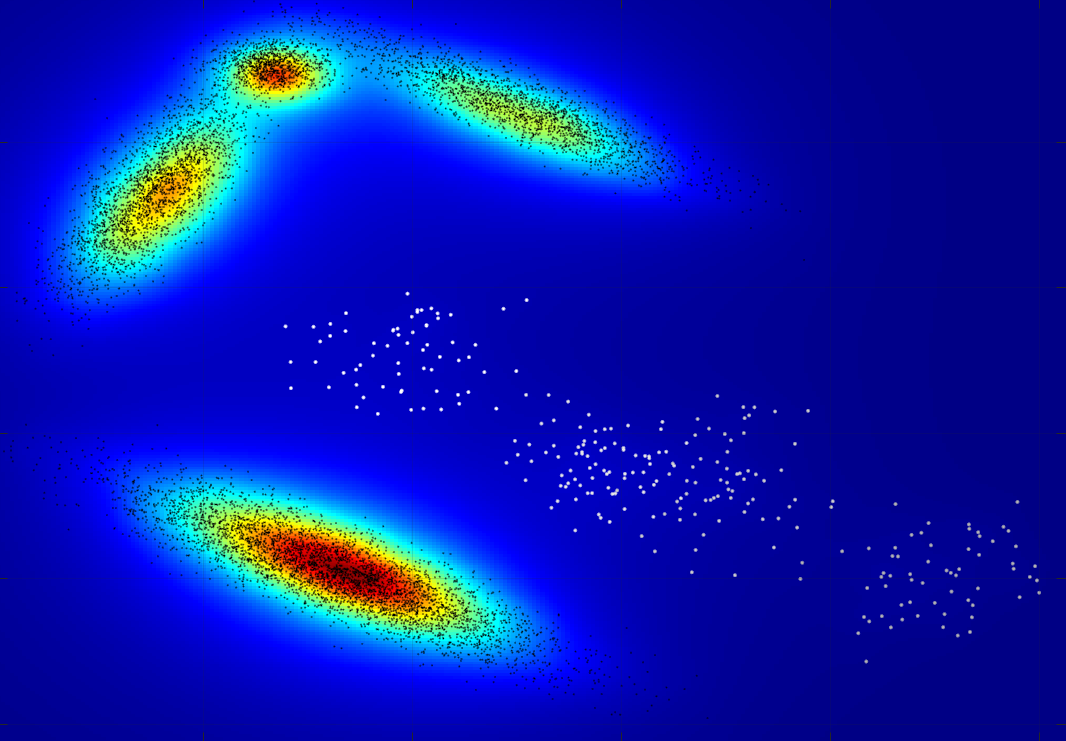}}
\vspace*{-6 mm}
\caption{Distribution of synthetic data.}
\label{syntheticData}
\end{wrapfigure}

\begin{figure}[htb]
\captionsetup[subfloat]{farskip=2pt,captionskip=1pt}
\centering
\subfloat[data]{
  \includegraphics[width=0.32\columnwidth]{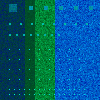}
}
\subfloat[RX]{
  \includegraphics[width=0.32\columnwidth]{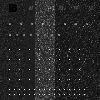}
}
\subfloat[KRX]{
  \includegraphics[width=0.32\columnwidth]{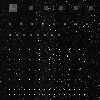}
}
\hspace{0mm}
\subfloat[targets]{
  \includegraphics[width=0.32\columnwidth]{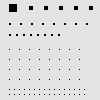}
}
\subfloat[RRX, D=3]{
  \includegraphics[width=0.32\columnwidth]{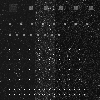}
}
\subfloat[RRX, D=50]{
  \includegraphics[width=0.32\columnwidth]{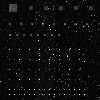}
}
\vspace*{-3 mm}
\caption{Detection responses for synthetic data}
\label{DetectionResponses}
\end{figure}

To compare the detectors in a controlled manner, we created a synthetic test image of size $100\times 100$ pixels with two spectral channels.
Fig. \ref{syntheticData} shows data distribution in a density colormap, here dark points are background and bright points are anomalies. Besides,  Fig.\ref{DetectionResponses}.a shows image as false colored (the first and the second dimensions assigned to the green and blue channels, respectively).
In this data, $2.72\%$ of the pixels are cast as anomalies (Fig. \ref{DetectionResponses}.d).

\vspace*{-4pt}

We added a small regularization value $\lambda=10^{-2}$ to better condition matrix inversions, and fixed the kernel lengthscale $\sigma$ as the median of all pairwise distances in both KRX and RRX. 
Detection responses are given in Fig. \ref{DetectionResponses} using whole image in training for RX and RRX, and randomly selected subset ($N=3000$) in training for KRX.
Note that background in synthetic data (cf. Fig. \ref{syntheticData}) is complex and non-Gaussian distributed.
For this data with complex background, RRX $D=3$ fails since 3 Fourier basis is not sufficient to model the background and  RX also fails since background is non-Gaussian distributed.
However, both RRX with only $50$ basis ($D=50$) and KRX ($N=3000$) are able to model background properly.

\begin{figure}[h]
\centering{\includegraphics[width=0.95\columnwidth]{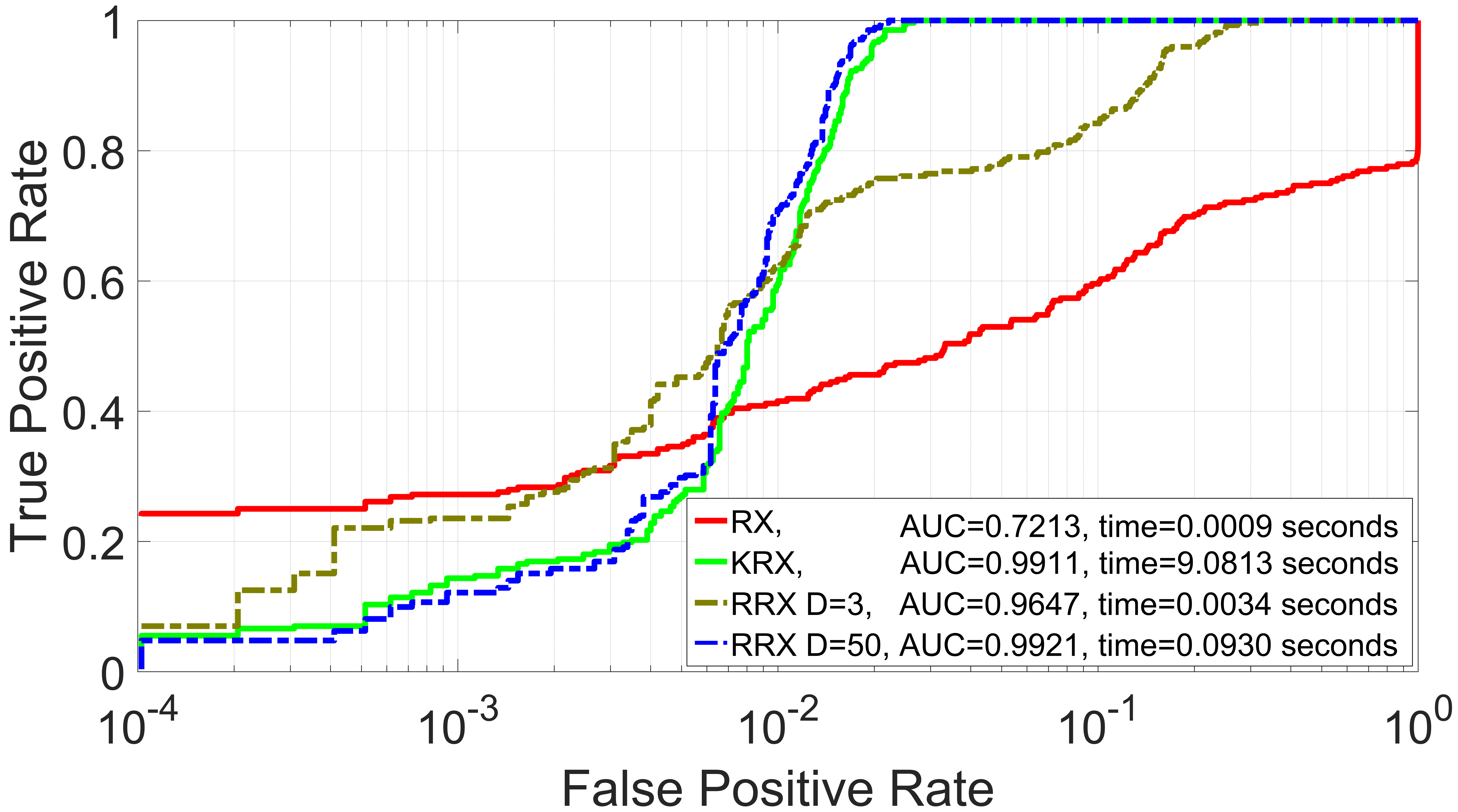}}
\vspace*{-3pt}
\caption{ROC curve for synthetic data.}
\label{syntheticDataROC}
\end{figure}

In Fig. \ref{syntheticDataROC} appears the ROC curves for all methods and they are given in log scale (abcissa) where area under the curve (AUC) values and execution times are given in legends. 
Since background is not Gaussian distributed, RX has the lowest performance.
KRX and the proposed RRX with $D=50$ basis show a similar superior performance, which is an expected result. 
However, computational load of KRX increases rapidly as the number of training samples increase.
In this paper, we used $N=3000$ randomly sampled pixels for KRX to achieve satisfactory detection results. 
The RRX detector with just $D=50$ basis has similar detection performance yet two orders of magnitude faster.

\subsection{Real-world dataset}
\label{sec:realworld_data_experiment}

\begin{figure}[h]
\centering{\includegraphics[width=0.95\columnwidth]{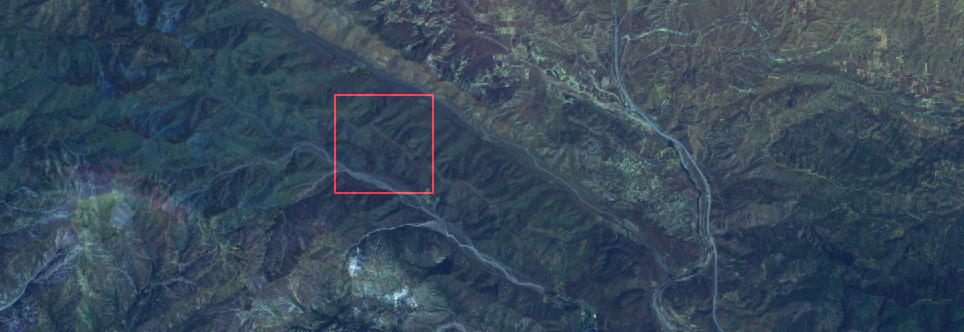}}
\vspace*{-4pt}
\caption{California dataset (3 channels are used for false color)}
\label{realworldData}
\end{figure}

For the real-world test, a multispectral image of California with 12 spectral channels and size of $964\times 332$ is used (see RGB composite in Fig. \ref{realworldData}).
California image is acquired from Sentinel-2 satellite at 8 August 2017 and downloaded from Google Earth Engine (GEE).

The image was divided into $16$ non-overlapping patches, each having size of $100\times 100$ pixels and anomalous pixels were synthetically fused into these patches using the mask given in Fig.\ref{DetectionResponses}.d. 
Half of the patches are used as a training set and the remaining half as the validation set.
The regularization parameter $\lambda$ was varied between $10^{-5}$ and $10^0$, and the lengthscale of the kernel was varied by taking values between $[0.05,5]$ and considering this particular values multiplied with $m$, where $m$ is the median of all pairwise distances. 
For each $\lambda$ and $c$ pairs, an average AUC value is calculated over the validation set with $8$ patches. 
The obtained optimal hyperparameters were $\lambda=10^{-4}$ and $c=0.5$. 

\begin{figure}[h]
\centering{\includegraphics[width=0.95\columnwidth]{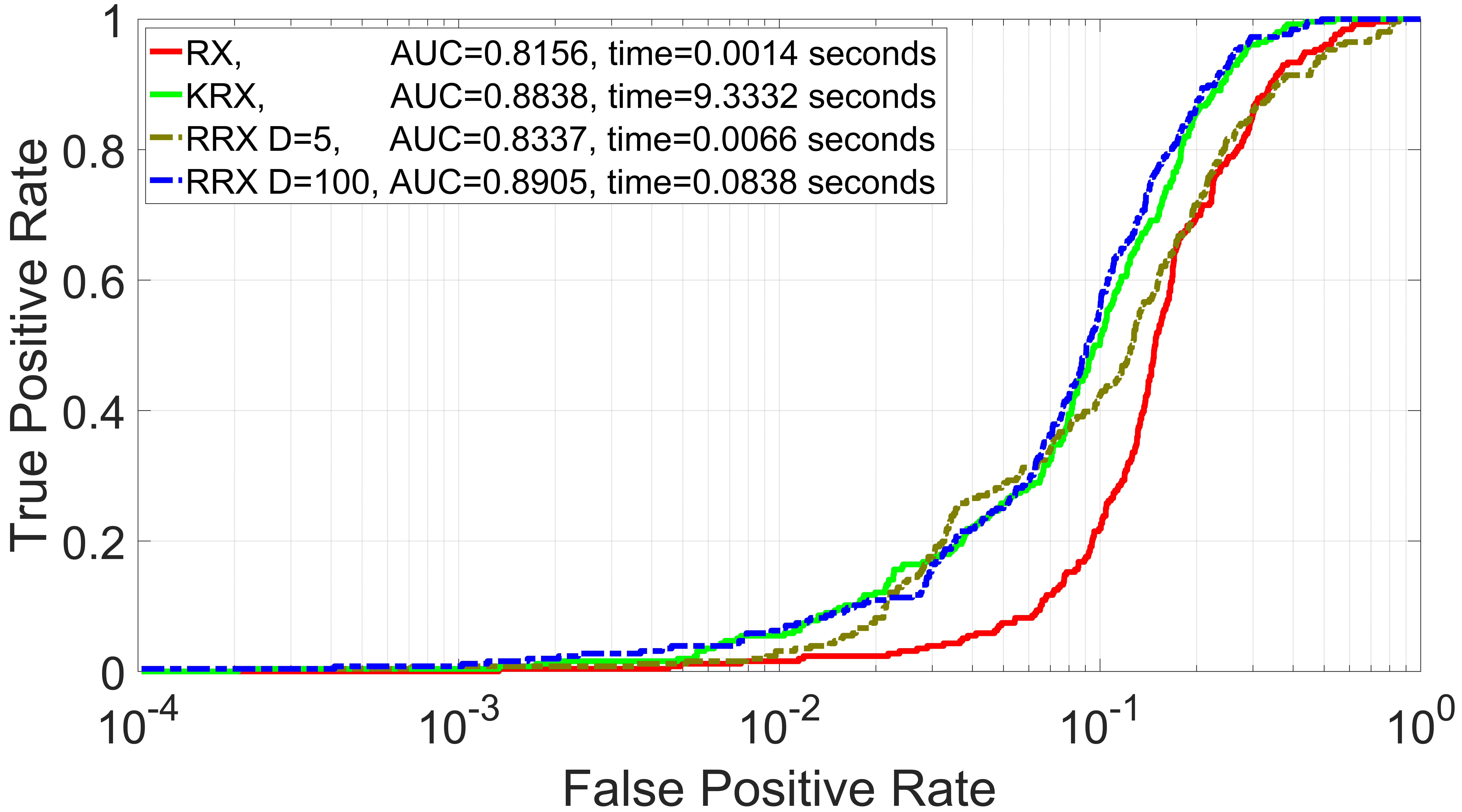}}
\vspace*{-5pt}
\caption{ROC curve for California data.}
\label{realworldROC}
\end{figure}

Test results are evaluated for the red square with size of $100\times 100$ which is shown in Fig. \ref{realworldData}.
Test results are given in Fig. \ref{realworldROC} in log scale where AUC values and execution times are given in legends. 
Proposed RRX can obtain similar performance with KRX using $D=100$ basis and hence it is two orders of magnitude faster for this data set.
Thus, better accuracies, robustness and computational efficiency is attained with the proposed RRX.

\section{Conclusions}
\label{sec:conclusion}

We introduced the use of random Fourier features in the context of anomaly detection. 
In particular, we focused on approximating the KRX method to cope with complex backgrounds in an computationally efficient manner. 
The proposed RRX provides space and time efficient solutions for anomaly detection, while providing high detection accuracy in complex nonlinear and non-Gaussian background distributions.
In our experiments, we observed that proposed RRX method can provide two orders of magnitude speedup compared to KRX method with no loss in accuracy.
Other kernelized detectors can benefit from this strategy, which will be the object of our future research.

\small
\bibliographystyle{IEEEbib}
\bibliography{bibFile}

\begin{thebibliography}{10}

\bibitem{Stein02}
D.~W.~J. Stein, S.~G. Beaven, L.~E. Hoff, E.~M. Winter, A.~P. Schaum, and A.~D.
  Stocker,
\newblock ``Anomaly detection from hyperspectral imagery,''
\newblock {\em IEEE Signal Processing Magazine}, vol. 19, no. 1, pp. 58--69,
  2002.

\bibitem{Keshava04}
N.~Keshava,
\newblock ``Distance metrics and band selection in hyperspectral processing
  with applications to material identification and spectral libraries,''
\newblock {\em IEEE Trans. Geosc. Rem. Sens.}, vol. 42, no. 7, pp. 1552--1565,
  2004.

\bibitem{Matteoli2010}
S.~Matteoli, M.~Diani, and G.~Corsini,
\newblock ``A tutorial overview of anomaly detection in hyperspectral images,''
\newblock {\em IEEE Aerospace and Electronic Systems Magazine}, vol. 25, no. 7,
  pp. 5--28, 2010.

\bibitem{Reed90tassp}
I.~S. Reed and X.~Yu,
\newblock ``Adaptive multiple-band {CFAR} detection of an optical pattern with
  unknown spectral distribution,''
\newblock {\em IEEE Transactions on Acoustics, Speech, and Signal Processing},
  vol. 38, no. 10, pp. 1760--1770, 1990.

\bibitem{Molero13jstars}
J.~M. Molero, E.~M. Garzón, I.~García, and A.~Plaza,
\newblock ``Analysis and optimizations of global and local versions of the {RX}
  algorithm for anomaly detection in hyperspectral data,''
\newblock {\em IEEE Journal of Selected Topics in Applied Earth Observations
  and Remote Sensing}, vol. 6, no. 2, pp. 801--814, 2013.

\bibitem{Zhao15rs}
Chunhui Zhao, Yulei Wang, Bin Qi, and Jia Wang,
\newblock ``Global and local real-time anomaly detectors for hyperspectral
  remote sensing imagery,''
\newblock {\em Remote Sensing}, vol. 7, no. 4, pp. 3966--3985, 2015.

\bibitem{Nar17dsp}
F.~Nar, E.~Okman, A.~Ozgur, and M.~Cetin,
\newblock ``Fast target detection in radar images using {R}ayleigh mixtures and
  summed area tables,''
\newblock {\em Digital Signal Processing}, vol. 77, pp. 86--101, 2018.

\bibitem{CampsValls09wiley}
G.~Camps-Valls and L.~Bruzzone, Eds.,
\newblock {\em Kernel methods for remote sensing data analysis},
\newblock Wiley \& Sons, UK, Dec 2009.

\bibitem{Kwon07}
H.~Kwon and N.~Nasrabadi,
\newblock ``A comparative analysis of kernel subspace target detectors for
  hyperspectral imagery,''
\newblock {\em {EURASIP} Journal of of Advances in Signal Proc.}, vol. 2007,
  no. 29250, 2007.

\bibitem{Longbotham14}
Nathan Longbotham and G.~Camps-Valls,
\newblock ``A family of kernel anomaly change detectors,''
\newblock in {\em IEEE Workshop on Hyperspectral Image and Signal Processing,
  Whispers}, June 2014.

\bibitem{Chen14taes}
S.~Y. Chen, Y.~Wang, C.~C. Wu, C.~Liu, and C.~I. Chang,
\newblock ``Real-time causal processing of anomaly detection for hyperspectral
  imagery,''
\newblock {\em IEEE Transactions on Aerospace and Electronic Systems}, vol. 50,
  no. 2, pp. 1511--1534, 2014.

\bibitem{Zhao16rs}
Chunhui Zhao, Xifeng Yao, and Bormin Huang,
\newblock ``Real-time anomaly detection based on a fast recursive kernel {RX}
  algorithm,''
\newblock {\em Remote Sensing}, vol. 8, no. 12, 2016.

\bibitem{Rahimi07nips}
Ali Rahimi and Benjamin Recht,
\newblock ``Random features for large-scale kernel machines,''
\newblock in {\em Advances in Neural Information Processing Systems 20}, pp.
  1177--1184. 2008.

\bibitem{Valero15igarss}
V.~Laparra, D.M. Gonzalez, D.~Tuia, and G.~Camps-Valls,
\newblock ``Large-scale random features for kernel regression,''
\newblock in {\em Geoscience and Remote Sensing Symposium (IGARSS), 2015 IEEE
  International}, 2015, pp. 17--20.

\bibitem{Perez17rks}
A.~Perez-Suay, J.~Amoros, L.~Gomez-Chova, V.~Laparra, Mu\ noz Mar\'i, and
  G.~Camps-Valls,
\newblock ``Randomized kernels for large scale earth observation
  applications,''
\newblock {\em Remote Sensing of Environment}, vol. 1, no. 1, pp. 1, 2017.

\bibitem{ShaweTaylor04}
John Shawe-Taylor and Nello Cristianini,
\newblock {\em Kernel Methods for Pattern Analysis},
\newblock {Cambridge University Press}, 2004.

\bibitem{Rahimi08}
Ali Rahimi and Benjamin Recht,
\newblock ``Weighted sums of random kitchen sinks: Replacing minimization with
  randomization in learning,''
\newblock in {\em Advances in Neural Information Processing Systems 21}, pp.
  1313--1320. Curran Associates, Inc., 2009.

\end{thebibliography}

\end{document}